\documentclass[conference]{IEEEtran}
\IEEEoverridecommandlockouts
\usepackage{cite}
\usepackage{array} 
\usepackage{adjustbox}

\usepackage{amsmath,amssymb,amsfonts}
\usepackage{algorithmic}
\usepackage{graphicx}
\usepackage{textcomp}
\usepackage{xcolor}
\usepackage{placeins}

\def\BibTeX{{\rm B\kern-.05em{\sc i\kern-.025em b}\kern-.08em
    T\kern-.1667em\lower.7ex\hbox{E}\kern-.125emX}}
\begin{document}

\title{NeuroPAL: Punctuated Anytime Learning with Neuroevolution for Macromanagement in Starcraft: Brood War\\
}

\author{\IEEEauthorblockN{Jim O'Connor}
\IEEEauthorblockA{\textit{Computer Science Department} \\
\textit{Connecticut College}\\
New London, CT, USA \\
joconno2@conncoll.edu}
\and
\IEEEauthorblockN{Yeonghun Lee}
\IEEEauthorblockA{\textit{Computer Science Department} \\
\textit{Connecticut College}\\
New London, CT, USA \\
ylee5@conncoll.edu}
\and
\IEEEauthorblockN{Gary B Parker}
\IEEEauthorblockA{\textit{Computer Science Department} \\
\textit{Connecticut College}\\
New London, CT, USA \\
parker@conncoll.edu}
}

\maketitle
\begin{abstract}
StarCraft: Brood War remains a challenging benchmark for artificial intelligence research, particularly in the domain of macromanagement, where long-term strategic planning is required. Traditional approaches to StarCraft AI rely on rule-based systems or supervised deep learning, both of which face limitations in adaptability and computational efficiency. In this work, we introduce NeuroPAL, a neuroevolutionary framework that integrates Neuroevolution of Augmenting Topologies (NEAT) with Punctuated Anytime Learning (PAL) to improve the efficiency of evolutionary training. By alternating between frequent, low-fidelity training and periodic, high-fidelity evaluations, PAL enhances the sample efficiency of NEAT, enabling agents to discover effective strategies in fewer training iterations. We evaluate NeuroPAL in a fixed-map, single-race scenario in StarCraft: Brood War and compare its performance to standard NEAT-based training. Our results show that PAL significantly accelerates the learning process, allowing the agent to reach competitive levels of play in approximately half the training time required by NEAT alone. Additionally, the evolved agents exhibit emergent behaviors such as proxy barracks placement and defensive building optimization, strategies commonly used by expert human players. These findings suggest that structured evaluation mechanisms like PAL can enhance the scalability and effectiveness of neuroevolution in complex real-time strategy environments.
\end{abstract}
\section{Introduction}
StarCraft: Brood War has long been a benchmark problem for artificial intelligence research, presenting challenges that extend beyond those seen in classical game AI domains like chess or Go. \cite{Buro2003RTSGA}\cite{surveyonGameAI} Released in 1998, Brood War remains relevant due to its combination of imperfect information, large state and action spaces, real-time decision-making, and multi-agent interactions. Unlike turn-based games, where each move can be deliberated extensively, StarCraft demands rapid responses to dynamically evolving situations, making it a compelling environment for testing AI methodologies \cite{RTSComp}.

Early AI approaches to StarCraft relied heavily on rule-based expert systems, scripting specific strategies that could execute well against predefined opponents \cite{synnaeve2011bayesian}. These methods, while effective in controlled scenarios, struggled to generalize against human opponents or adapt to novel strategies. More recently, machine learning techniques have been applied to StarCraft, particularly in the context of deep learning \cite{justesen2017learningmacromanagementstarcraftreplays} and decision trees \cite{decisiontreeBroodwar}. Notable works include Justesen and Risi’s macromanagement learning from replays, which used deep learning to extract build-order decisions from human games. These methods have demonstrated the viability of neural networks and deep learning in RTS decision-making but remain computationally expensive and heavily reliant on supervised pretraining. Additionally, existing methods rely heavily on action selection between pre-scripted strategies to remain viable. \cite{tang2018reinforcement} \cite{xu2019macroactionselectiondeep} \cite{zhang2020reinforcementStrategy} \cite{wu2023mscdatasetmacromanagementstarcraft}

StarCraft II, the successor to Brood War, has been more thoroughly explored by AI researchers, culminating in AlphaStar achieving Grandmaster-level play \cite{alphastar}. However, StarCraft II differs in key ways that make it more tractable for AI: it features a structured API that provides cleaner state information,  and the game itself removes many execution challenges through action abstraction. Brood War, by contrast, presents a far more challenging learning landscape due to its lower-level API, requiring agents to handle unit control at a granular level, and its more technically demanding and unforgiving gameplay. As a result, Brood War remains unsolved and continues to serve as a proving ground for AI techniques that must operate under greater constraints.

To address these challenges, we introduce NeuroPAL, an evolutionary approach that combines Neuroevolution of Augmenting Topologies (NEAT) with Punctuated Anytime Learning (PAL). NEAT has previously been used to evolve neural networks for StarCraft micro-management tasks, \cite{zhen2013neuroevolution} leveraging its ability to optimize both network topology and weights over generations. PAL, developed by Parker \cite{parker1999adaptive}, introduces a punctuated learning approach that balances frequent low-fidelity training with periodic high-fidelity evaluations. By integrating these methods, NeuroPAL enables efficient evolutionary training, alternating between rapid iterations with limited simulations and punctuated evaluations with extensive testing to ensure robust strategy development. This approach mitigates issues related to sample inefficiency and speeds up training while preserving the adaptability of evolutionary learning in complex environments.

\section{Starcraft}
StarCraft: Brood War (SC:BW) is a real-time strategy (RTS) game developed by Blizzard Entertainment, first released in 1998 \cite{starcraft}. Despite its age, SC:BW remains one of the most complex and strategically rich RTS games ever made, with a dedicated player base and an enduring presence in esports. Unlike many modern RTS games, SC:BW lacks features such as automated unit control and unlimited selection groups, requiring players and AI agents to manage a variety of tasks manually. This results in a high skill ceiling, making it a compelling challenge for artificial intelligence research.

SC: BW's complexity comes from its multi-layered strategic and mechanical depth. Players must manage economic resources, construct buildings, and control units in real time while responding dynamically to their opponent’s actions. The game features three distinct races, each with unique units, abilities, and playstyles, further increasing the complexity of strategic decision-making. The necessity for both micro-level unit control and macro-level strategic planning presents a significant challenge to AI, requiring agents to handle long-term planning and tactical execution simultaneously.

\begin{figure}
    \centering
    \includegraphics[width=1\linewidth]{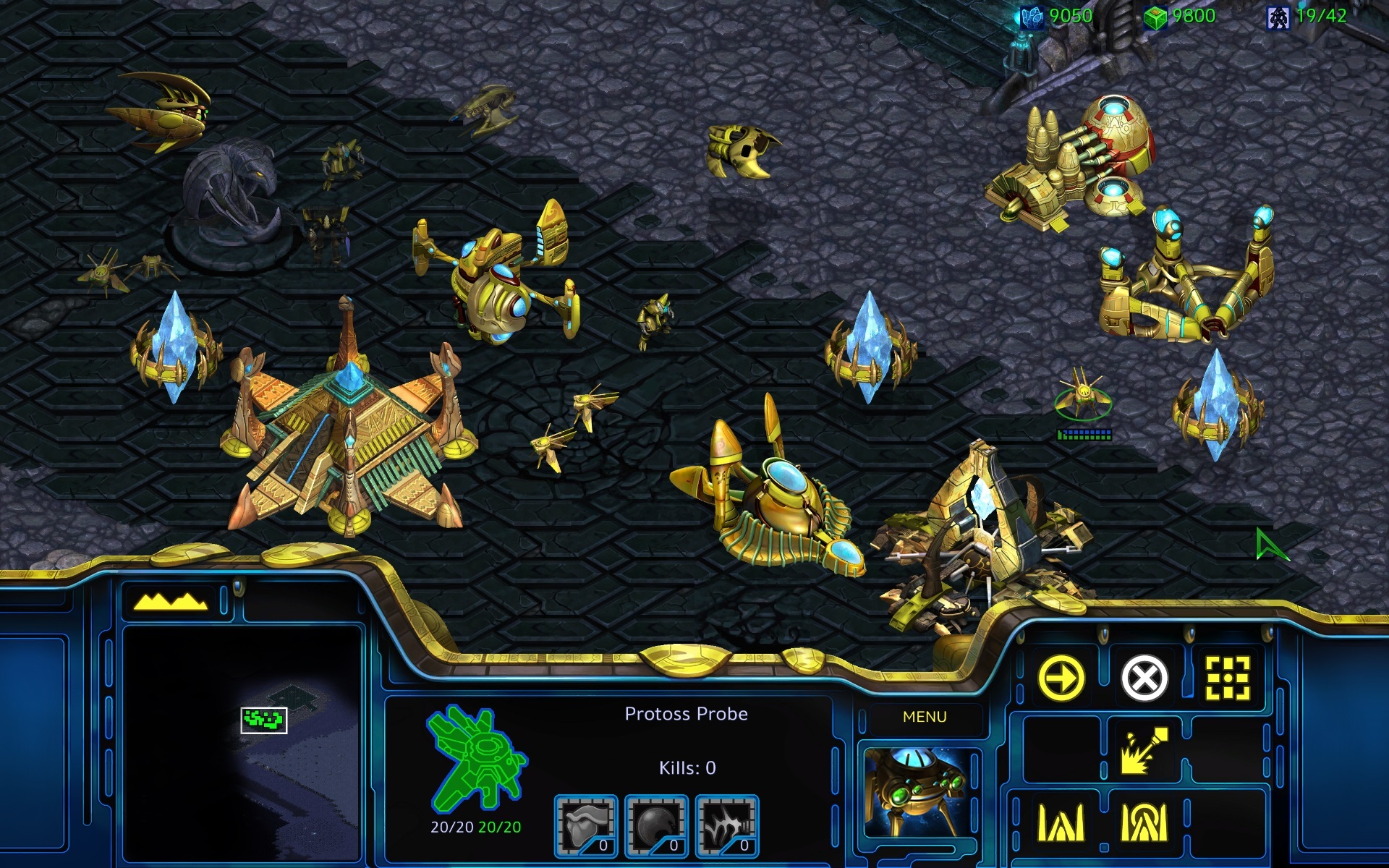}
    \caption{Screenshot of SC:BW while playing as Protoss. SC:BW is a real-time strategy game where each players produce buildings and army units to defeat other players.}
    \label{starcraft}
\end{figure}

Recognizing the potential of SC:BW as an AI benchmark and challenge, tools have been built to facilitate AI research. The Brood War API (BWAPI), \cite{DavidChurchill} is the most widely used interface for integrating AI with the SC:BW engine, allowing researchers to develop autonomous agents that interact directly with the game. BWAPI provides real-time access to game state information and enables programmatic control of units, but it does not grant omniscient access or full automation, preserving many of the game's inherent challenges. This makes it an ideal environment for building and evaluating intelligent agents that have to overcome a wide range of complex challenges.

AI research in SC:BW has traditionally focused on two main areas: micromanagement, which involves optimizing unit control and combat efficiency, and macromanagement, which encompasses economy construction, build order execution, and high-level strategic planning. While micromanagement-focused approaches such as reinforcement learning for unit control \cite{shao2018starcraftmicromanagementreinforcementlearning} have demonstrated success in limited scenarios, macromanagement remains a significantly harder problem. Unlike StarCraft II, which has seen breakthroughs with deep reinforcement learning methods, SC:BW presents additional obstacles such as more restrictive unit controls, and a lack of structured state representations. As a result, SC:BW remains comparably unsolved, making it an exciting and compelling research platform.

Our approach, NeuroPAL, tackles the challenge of macromanagement by integrating neuroevolution with an adaptive learning schedule leveraging Punctuated Anytime Learning. Using BWAPI, we develop agents that evolve strategic decision-making over successive generations, alternating between rapid iterative improvement phases and in-depth evaluation periods, resulting in a neuroevolution system with a significantly increased learning efficiency.

\section{Neuroevolution of Augmenting Topologies}

NeuroEvolution of Augmenting Topologies (NEAT) is an evolutionary algorithm designed for optimizing neural networks. Developed by Kenneth Stanley in the early 2000s, \cite{NEAT} NEAT introduced several key innovations that distinguish it from prior neuroevolution techniques. Traditional evolutionary algorithms for neural networks typically evolve only the connection weights of a fixed-topology network, which limits their ability to discover novel architectures. NEAT, in contrast, evolves both the structure and weights of neural networks simultaneously, enabling the generation of more complex and effective topologies over generations. This is achieved through a genetic encoding scheme that preserves innovation via historical markings, allowing structurally different networks to compete fairly. The algorithm employs speciation to maintain diversity within the evolving population, preventing the dominance of a single topology and allowing multiple approaches to be explored structurally in parallel. These features have made NEAT a widely studied and influential method in evolutionary computation and neural network optimization \cite{NEATReview}.

NEAT has been widely applied in AI research for its ability to evolve adaptable and high-performing policies without requiring extensive pretraining.\cite{NEATReview2} It has been particularly effective in domains where handcrafted features are impractical, and reinforcement learning struggles with sparse rewards. In video game AI, NEAT has demonstrated success in learning control policies for various environments, including racing games, strategy games, and Atari.

Within the RTS genre, NEAT has been explored for tasks such as unit micro-management, build-order optimization, and high-level decision-making. Games like StarCraft: Brood War pose significant challenges due to their real-time nature, vast state space, and the need for long-term planning. NEAT’s ability to incrementally evolve solutions makes it particularly well-suited for such complex environments.

\section{Punctuated Anytime Learning}

Punctuated Anytime Learning (PAL) is an AI learning framework designed to integrate real-time adaptation with evolutionary computation. The concept builds upon the idea of anytime learning, where a system continuously refines its solution while operating, allowing for incremental improvements over time. PAL enhances this process by introducing periodic evaluations on the real environment among simulations, effectively punctuating the learning phase with moments of real-world validation. This methodology addresses the common challenge in evolutionary learning where models trained exclusively in simulation often fail to generalize well to real-world conditions, often referred to as the sim2real gap \cite{jaunet2021sim2realviz}.

PAL has been successfully applied in various domains, including robotics \cite{Parker2009} and game AI \cite{Parker2015}. In robotics, it has been used to evolve locomotion strategies where an initial phase of simulation-based learning is periodically refined through real-world testing, ensuring adaptability to unexpected physical constraints \cite{Parker2009}. Similarly, in video game AI, PAL has been employed to evolve competitive agents that learn from simulated gameplay but periodically test strategies against human or AI opponents, refining their decision-making processes in response to actual gameplay dynamics \cite{Parker2015}.

\begin{figure}
    \centering
    \includegraphics[width=0.9\linewidth]{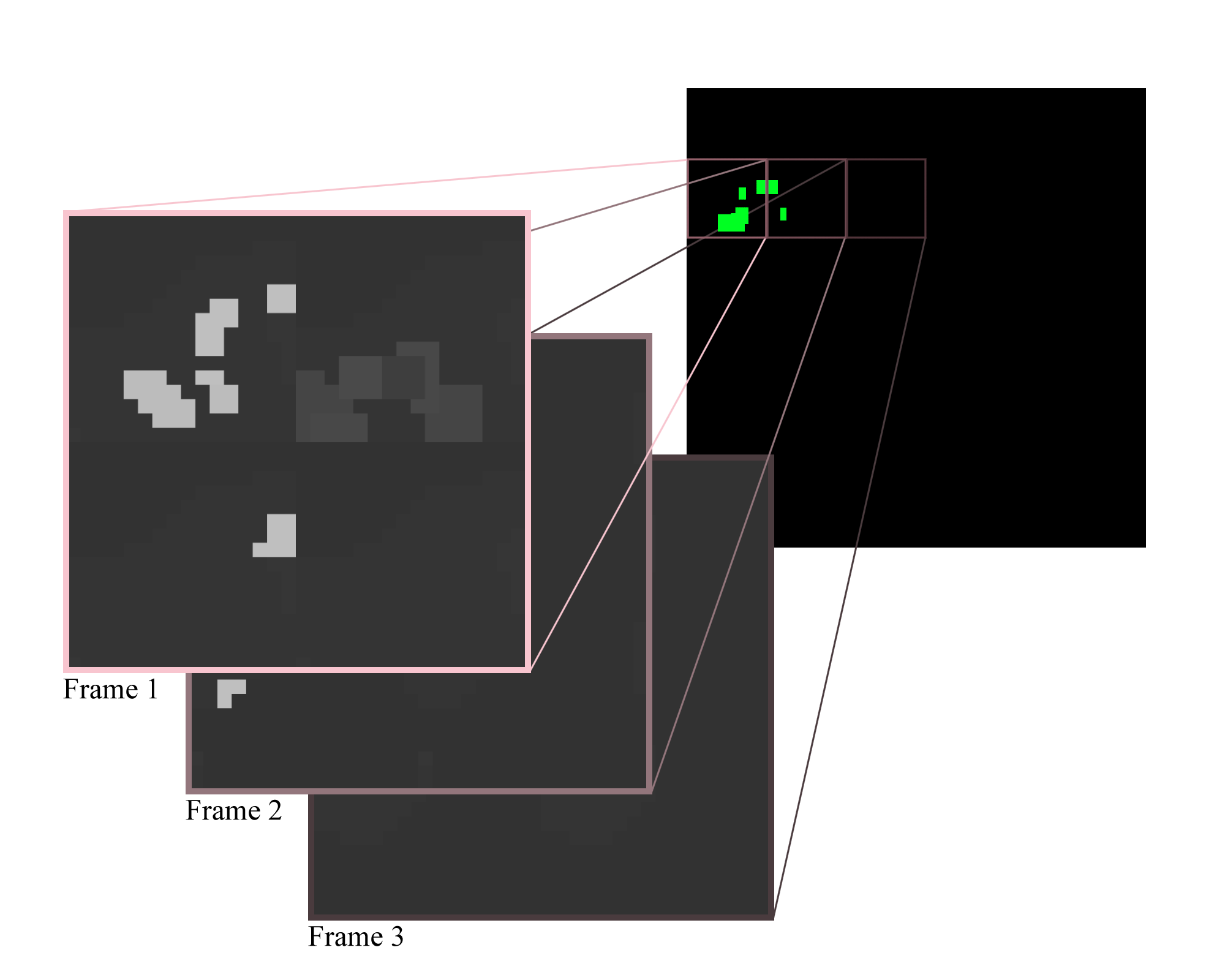}
    \caption{Inputs are 32x32 subsections of a map at each frame. The input is parsed from 1 subsection per frame.}
    \label{agent-input-1}
\end{figure}

Within real-time strategy (RTS) games, we believe PAL has promise in addressing challenges of strategic adaptation and long-term planning. Given the complexity of RTS environments, where decisions must balance short-term tactical advantages with long-term resource management, PAL provides a structured way to refine AI strategies dynamically. By punctuating evolutionary training evaluation based on quick single-game evaluations with periods of longer and, therefore, more accurate evaluations, PAL enables the AI to adjust to complex game conditions more effectively than purely simulated approaches. This capability also makes PAL an ideal candidate for integration with neuroevolution techniques like NEAT, where evolving neural architectures benefit from iterative refinement against real opponents.

In the context of StarCraft: Brood War, PAL facilitates the learning of complex strategies by allowing the AI to iterate quickly through many candidate solutions and adjust its policies based on a robust evaluation of match performance. The combination of PAL with NEAT enables the evolution of increasingly sophisticated strategies that balance economic expansion, unit production, and tactical engagements.

\section{Neuro-PAL}
NeuroPAL combines NEAT with Punctuated Anytime Learning (PAL) to enhance the efficiency of evolutionary training for macromanagement in StarCraft: Brood War. By integrating NEAT’s ability to evolve both network topology and weights with PAL’s structured evaluation mechanism, NeuroPAL accelerates learning while maintaining strategic adaptability. Traditional evolutionary training often suffers from slow convergence and inefficient sample utilization, particularly in complex environments such as SC:BW, where long-term planning is crucial. By leveraging PAL’s periodic high-fidelity evaluations, NeuroPAL mitigates these challenges, allowing for more effective strategic evolution over successive generations.

\begin{figure}
    \centering
    \includegraphics[width=0.9\linewidth]{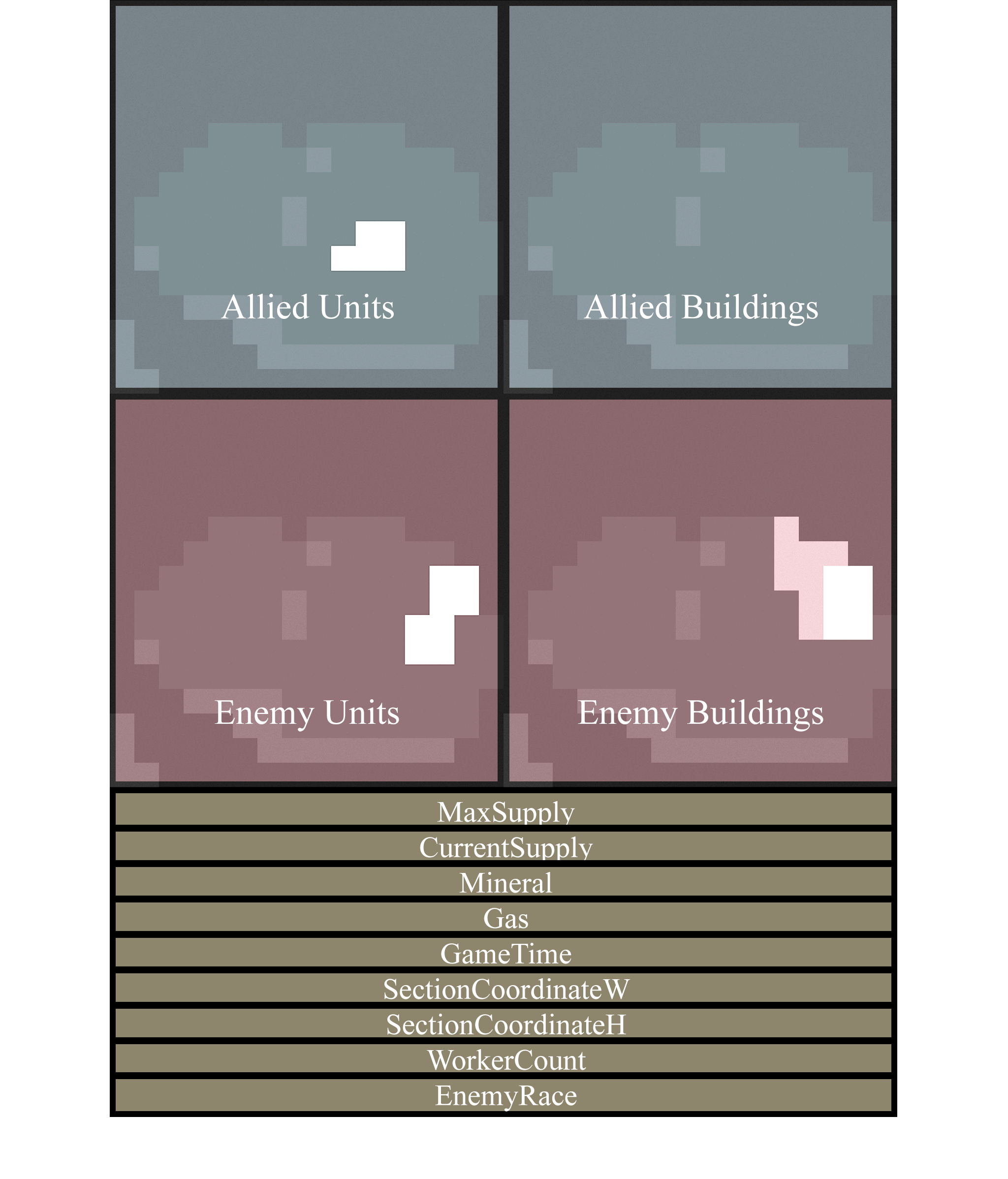}
    \caption{The graphical representation of the input matrix after parsing the 32x32 subsection. Allied Units, Allied Buildings, Enemy Units, and Enemy Buildings are grayscale images colored here for easier interpretation. The grayscale images represent the placement of tile types, units, and buildings within the subsection.}
    \label{agent-input-2}
\end{figure}

\subsection{Fitness Biasing and Punctuated Anytime Learning}

PAL operates by alternating between frequent, low-fidelity evaluations and less frequent, high-fidelity evaluations. In the context of NeuroPAL, fitness biasing \cite{Parker2015} \cite{parker2011fitness} is employed to adjust fitness values based on results from the most recent high-fidelity evaluation. This mechanism ensures that high-performing chromosomes from punctuated generations retain their advantage even in subsequent rapid training cycles, thereby improving convergence efficiency. The weighted fitness function is defined as follows:

\[f_{weighted} = f_{raw}\times \frac{f_{pi}} {max(f_p)}\]

$f_{raw}$ represents the immediate fitness score of a chromosome, and $f_{pi}$ corresponds to the fitness recorded during the most recent punctuated evaluation. This approach effectively prevents premature convergence toward suboptimal strategies by preserving diversity and ensuring that successful strategies identified during high-fidelity evaluations are propagated efficiently.

Specifically in the architecture of our NeuroPAL system, the algorithm evaluates 15 standard generations before evaluating 1 punctuated or high-fidelity evaluation. In this punctuated generation, the results of 50 games are averaged together to produce a single more robust score, as opposed to the 10-game score that is used in non-punctuated generations.

SC:BW is a complex environment where a single evaluation does not fully capture the efficacy of the agent due to the variance of the opponent's starting positions, race, and strategy. To partially limit the variance of the environment, we had our agent play as the Terran race on the map (2)Benzene against an in-game agent who also played Terran. 

\subsection{Action Representation and Decision-Making}
To facilitate strategic decision-making, NeuroPAL structures the input and output representations to align with macromanagement tasks. Input features include a 32x32 subsection of the game map, unit compositions, economic statistics, and opponent race information. The network outputs an action vector encoding unit production decisions, macromanagement directives, and locational parameters for strategic execution. These outputs are parsed through an abstraction layer adapted from the well-known agent SteamHammer \cite{Steamhammer}, ensuring compatibility with the SC:BW engine while enabling adaptive strategic behavior.

More specifically, our agent loads an area 32 tiles wide and 32 tiles high for each frame during a match. This 32x32 subsection is copied into 4 further 32x32 images, each of which is categorized to a unit type: either allied units, allied buildings, enemy units, or enemy buildings, as seen in Figure 3. These 32x32 images are then max-pooled with either a BuildScore or KillScore value from BWAPI to reduce the size of the images to 16x16 where they are then recombined into a final 32x32 input image. The image is populated with integers representing tile types (FOG, Walkable, NotWalkable) and unit types (Terran\_Marine, Terran\_Command\_Center). Nine additional rows of gameplay values are also added to the bottom of the input matrix as described in Fig. 3. This additional information is used to balance the object representation of the environment with the image representation of the environment. The total input size of the model is 1312.

The output space of the model consists of 70 seperate values. The first 56 outputs correspond to all types of possible units and upgrades available in-game to the Terran race. The next 12 outputs consists of macro-commands that mirror the internal action types used by SteamHammer. The last 2 outputs are continuous values between 0 and 1 that represent the X and Y coordinates in the map as ratios of the map size. This location-based value represents where the macro action that is selected will be executed on the map. The action space is then parsed into actions used by the abstraction layer from the SteamHammer bot. This architecture helps to limit the complexity of the environment to purely macromanagement. All possible actions the agent can make is described in Table I.

\begin{table}[htbp]
    \caption{All possible actions the NEAT/NeuroPAL agent can take. For actions that require a specified coordinate, the agent uses two continuous value outputs as x and y coordinate outputs.}
    \centering 
    \begin{tabular}{|c|p{60mm}|} 
        \hline
        \textbf{Action Types} & \textbf{Description} \\
        \hline
        None & No Action\\
        \hline
        Scout & The agent sends a worker unit to the designated coordinate. \\
        \hline
        StartGas & The agent constructs a Refinery on the closest vespine geyser from the designated coordinate and begins mining.\\
        \hline
        StopGas & The agent reassigns workers that were mining gas from the closest Refinery from the coordinate to mine minerals.\\
        \hline
        Aggressive & The agent sends out its army units to attack the enemy base. A small part of the army is sent to search for enemy buildings if there are no known enemy buildings left in enemy spawn location.\\
        \hline
        Defensive & The agent places its army units at its spawn, defending the base from incoming attacks.\\
        \hline
        PullWorkers & Worker units gathering resources will stop gathering resources and start attacking nearby enemy units.\\
        \hline
        ReleaseWorkers & Worker units around the specified coordinates will stop mining from that location.\\
        \hline
        PostWorker & The agent posts the closest free worker to the given tile location.\\
        \hline
        UnPostWorker & The agent releases all workers within a 5 tile radius from post\\
        \hline
        Scan & If a ComSat Addon building is constructed, the agent will use a Scan on the specified coordinate. \\
        \hline
        Spidermine & If Vultures upgraded with Spidermines exist, the agent will order nearby vultures to place spidermines at the specified location.\\
        \hline
    \end{tabular}
    \label{agent-action-types}

\end{table}

\subsection{Fitness Function}
The fitness function of the agent incorporates several measures, such as the efficacy of the type of unit produced, enemy unit destruction, and enemy building destruction. \(B_c\) represents the number of buildings constructed, \(U_k\) is the number of enemy army units killed, and \(W\) is a victory value. Enemy buildings destroyed \(B_k\) and army unit production indicate that the agent is engaging in active combat that could potentially lead to victory, both of which provide a clearer view of the efficacy of the agent in the system.

\[f_{raw} = 10B_c + 100(U_k+W) + 1000(B_k + U_p) + \frac{500A_A}{A_T}\]

This raw fitness value is used to calculate the immediate fitness of a chromosome and is a part of our weighted fitness calculation discussed earlier. 

\subsection{Mitigating bias towards immediate fitness gains}
A common challenge in evolutionary training is the tendency for agents to prioritize immediate fitness gains over long-term strategic advantages. In SC:BW, early-game actions such as constructing a Command Center provide immediate fitness rewards, whereas higher-tier strategies require incremental setup and investment. To address this, NeuroPAL incorporates an output space availability score, calculated as:

\[S_{availability} = \frac{A_A}{A_T}\]

$A_A$ represents the number of available actions, and $A_T$ denotes the total potential actions. By rewarding strategies that expand the available action space, this approach encourages the agent to explore more sophisticated build orders and strategic planning.

\section{Results}
Through repeated experimentation, we observed that NeuroPAL significantly improved training efficiency compared to NEAT alone. By alternating between low-fidelity rapid training and high-fidelity punctuated evaluations, NeuroPAL achieved higher fitness much more quickly than traditional NEAT-based approaches.

In our experiments, NeuroPAL consistently identified viable strategies earlier than NEAT-only training. For instance, within the first 20,000 evaluations, NeuroPAL-based agents had begun exhibiting effective behavior and increases in fitness, whereas NEAT-only training required at least 40,000 or more evaluations to reach comparable performance. This accelerated learning process is particularly beneficial in computationally expensive environments like SC:BW, where training iterations are time-intensive.

\begin{figure}
    \centering
    \includegraphics[width=1\linewidth]{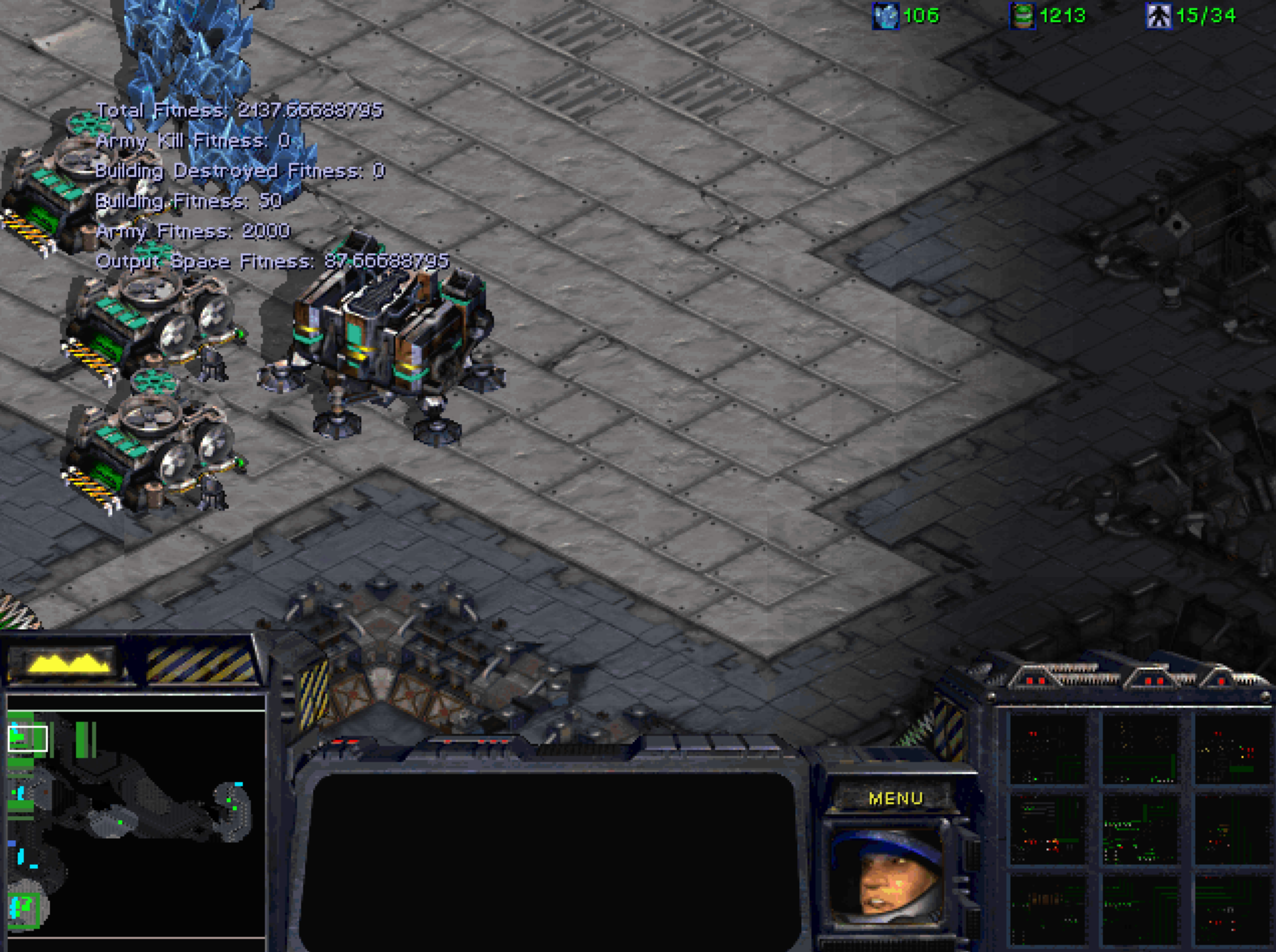}
    \caption{A Barrack and 3 Supply Depots are placed in the corner of the map.}
    \label{emergent-behavior-1}
\end{figure}

\begin{figure}
    \centering
    \includegraphics[width=1\linewidth]{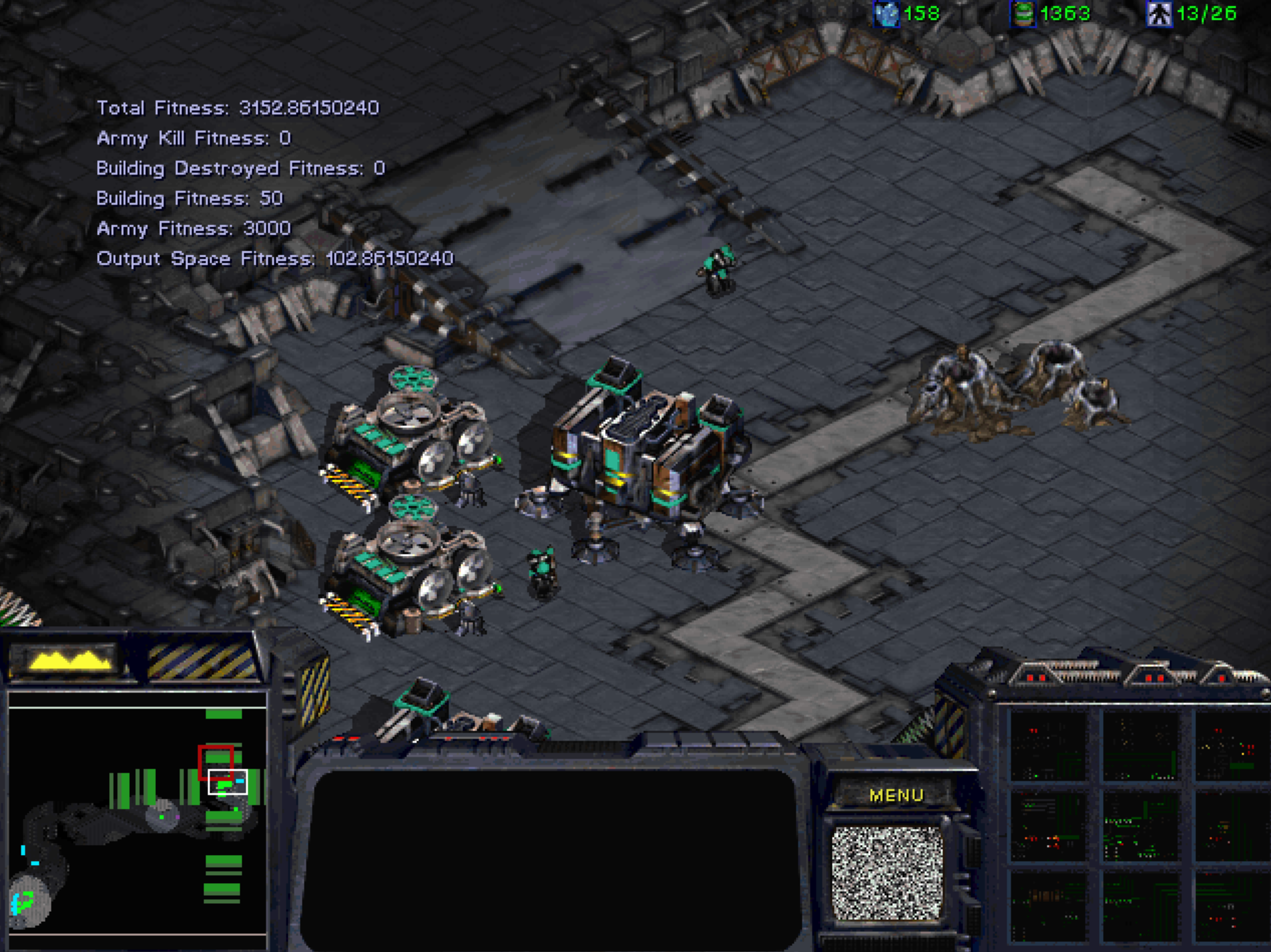}
    \caption{2 Barracks and 2 Supply Depots are placed near the ramp of the enemy base. This is a prominent example of the Proxy Barrack Strategy.}
    \label{emergent-behavior-2}
\end{figure}

\subsection{Emergent Behavior}

Another key finding that we observed was the strategic placement of buildings to maximize defensive efficiency. Initially, buildings were constructed haphazardly, with little regard for terrain features or vulnerability to enemy attacks. Over time, however, the agents began employing placement strategies that minimized vulnerability while optimizing accessibility.

For instance, agents learned to position Barracks near natural choke points, creating an obstacle that limited enemy mobility while ensuring a shorter travel distance for reinforcements. Additionally, Supply Depots were frequently positioned in defensive clusters, forming walls that restricted enemy movement and provided protection to vulnerable economic structures. Figure 4, for example, shows a defensive cluster of buildings that prevent attacks around the agent's minerals. This behavior aligns with human strategies commonly employed in high-level play, where building placement is used to create defensive advantages.

A particularly striking example of emergent strategy was the occasional use of “proxy Barracks” tactics, where agents placed production buildings near enemy bases to enable early aggression, as shown in Figure 5. This strategy, commonly used in professional play, allows for rapid unit deployment, catching opponents off guard before they can adequately defend. The fact that such tactics arose purely from evolutionary optimization suggests that NeuroPAL can demonstrate interesting examples of strategic decision-making in StarCraft.
\flushbottom 
\begin{figure*}
    \centering
    \includegraphics[width=0.7\linewidth]{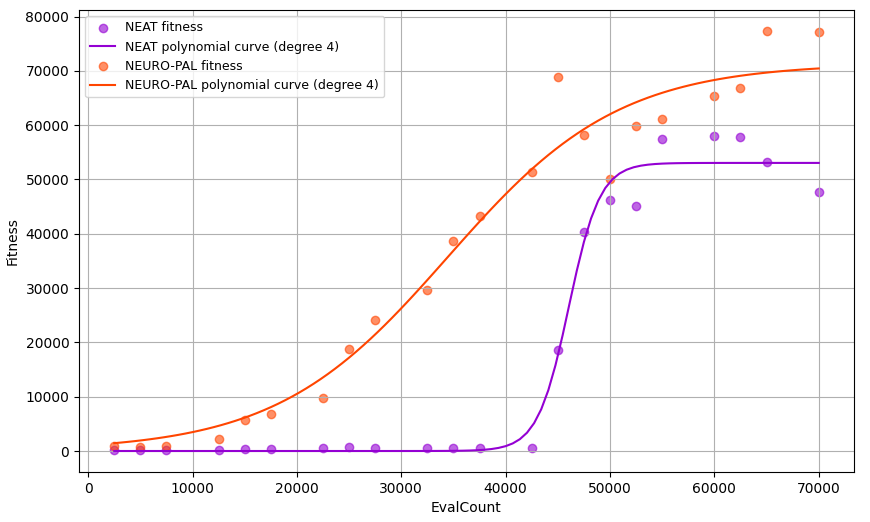}
    \caption{This graph compares 10 training runs on NeuroPAL against a NEAT training run. NeuroPAL demonstrates a rapid increase in fitness while also achieving similar fitness to NEAT at a 20\% lower number of evaluations.}
    \label{fig:enter-label}
\end{figure*}
\FloatBarrier
\subsection{Learning Efficiency}

The structured evaluation cycles introduced by PAL significantly accelerated the emergence of these behaviors. Compared to standard NEAT-based training, which often required tens of thousands of evaluations to develop competitive strategies, NeuroPAL-trained agents demonstrated sophisticated macromanagement behaviors within the first 20,000 evaluations [Fig. 6]. The high-fidelity evaluation phases played a crucial role in reinforcing successful strategies, ensuring that advantageous behaviors persist across generations.

Overall, our findings demonstrate that PAL-enhanced neuroevolution not only improves learning efficiency but also fosters the emergence of high-level strategic behavior. The NeuroPAL's ability to autonomously discover competitive strategies highlights its potential for advancing AI-driven decision-making in complex real-time strategy environments.

\section{Conclusion}
NeuroPAL successfully integrates NEAT with PAL to facilitate the evolution of strategic macromanagement in StarCraft: Brood War. This work's primary contribution is demonstrating that PAL enhances the efficiency of evolutionary training, allowing agents to reach competitive performance in fewer generations. By structuring the evaluation process into alternating rapid-learning and high-fidelity assessment phases, PAL ensures that high-performing strategies are reinforced earlier, mitigating inefficiencies inherent in traditional neuroevolutionary approaches.

In addition, the emergence of well-established StarCraft strategies, such as proxy barracks and optimized defensive structures, underscores the capability of neuroevolution to develop competitive play without direct human supervision. These results suggest that structured evaluation mechanisms like PAL can serve as a crucial enhancement to neuroevolutionary techniques in complex, real-time environments.

\bibliographystyle{ieeetr}
\bibliography{bib}

\begin{thebibliography}{10}

\bibitem{Buro2003RTSGA}
M.~Buro and T.~Furtak, ``Rts games and real-time ai research,'' 2003.

\bibitem{surveyonGameAI}
S.~Yildirim~Yayilgan and S.~Stene, {\em A Survey on the Need and Use of AI in Game Agents}.
\newblock 03 2010.

\bibitem{RTSComp}
M.~Buro and D.~Churchill, ``Real-time strategy game competitions,'' {\em AI Magazine}, vol.~33, no.~3, pp.~106--108, 2012.

\bibitem{synnaeve2011bayesian}
G.~Synnaeve and P.~Bessiere, ``A bayesian model for opening prediction in rts games with application to starcraft,'' in {\em 2011 IEEE Conference on Computational Intelligence and Games (CIG'11)}, pp.~281--288, IEEE, 2011.

\bibitem{justesen2017learningmacromanagementstarcraftreplays}
N.~Justesen and S.~Risi, ``Learning macromanagement in starcraft from replays using deep learning,'' 2017.

\bibitem{decisiontreeBroodwar}
I.~Zelinka and L.~Sikora, ``Starcraft: Brood war — strategy powered by the soma swarm algorithm,'' in {\em 2015 IEEE Conference on Computational Intelligence and Games (CIG)}, pp.~511--516, 2015.

\bibitem{tang2018reinforcement}
Z.~Tang, D.~Zhao, Y.~Zhu, and P.~Guo, ``Reinforcement learning for build-order production in starcraft ii,'' in {\em 2018 Eighth International Conference on Information Science and Technology (ICIST)}, pp.~153--158, IEEE, 2018.

\bibitem{xu2019macroactionselectiondeep}
S.~Xu, H.~Kuang, Z.~Zhuang, R.~Hu, Y.~Liu, and H.~Sun, ``Macro action selection with deep reinforcement learning in starcraft,'' 2019.

\bibitem{zhang2020reinforcementStrategy}
T.~Zhang, X.~Li, X.~Li, G.~Liu, and M.~Tian, ``Reinforcement learning based strategy selection in starcraft: Brood war,'' in {\em Proceedings of the 2020 Artificial Intelligence and Complex Systems Conference}, pp.~121--128, 2020.

\bibitem{wu2023mscdatasetmacromanagementstarcraft}
H.~Wu, Y.~Zong, J.~Zhang, and K.~Huang, ``Msc: A dataset for macro-management in starcraft ii,'' 2023.

\bibitem{alphastar}
O.~Vinyals, I.~Babuschkin, W.~M. Czarnecki, M.~Mathieu, A.~Dudzik, J.~Chung, D.~H. Choi, R.~Powell, T.~Ewalds, P.~Georgiev, {\em et~al.}, ``Grandmaster level in starcraft ii using multi-agent reinforcement learning,'' {\em nature}, vol.~575, no.~7782, pp.~350--354, 2019.

\bibitem{zhen2013neuroevolution}
J.~S. Zhen and I.~Watson, ``Neuroevolution for micromanagement in the real-time strategy game starcraft: Brood war,'' in {\em AI 2013: Advances in Artificial Intelligence: 26th Australasian Joint Conference, Dunedin, New Zealand, December 1-6, 2013. Proceedings 26}, pp.~259--270, Springer, 2013.

\bibitem{parker1999adaptive}
G.~B. Parker and J.~W. Mills, ``Adaptive hexapod gait control using anytime learning with fitness biasing,'' in {\em Proceedings of the 1st Annual Conference on Genetic and Evolutionary Computation-Volume 1}, pp.~519--524, 1999.

\bibitem{starcraft}
``Starcraft: Brood war,'' Blizzard Entertainment and Saffire Entertainment, videogame, 1998, available on Windows and Mac OS.

\bibitem{DavidChurchill}
S.~Ontañón, G.~Synnaeve, A.~Uriarte, F.~Richoux, D.~Churchill, and M.~Preuss, ``A survey of real-time strategy game ai research and competition in starcraft,'' {\em IEEE Transactions on Computational Intelligence and AI in Games}, vol.~5, no.~4, pp.~293--311, 2013.

\bibitem{shao2018starcraftmicromanagementreinforcementlearning}
K.~Shao, Y.~Zhu, and D.~Zhao, ``Starcraft micromanagement with reinforcement learning and curriculum transfer learning,'' 2018.

\bibitem{NEAT}
K.~O. Stanley and R.~Miikkulainen, ``Evolving neural networks through augmenting topologies,'' {\em Evolutionary Computation}, vol.~10, no.~2, pp.~99--127, 2002.

\bibitem{NEATReview}
E.~Galván and P.~Mooney, ``Neuroevolution in deep neural networks: Current trends and future challenges,'' {\em IEEE Transactions on Artificial Intelligence}, vol.~2, no.~6, pp.~476--493, 2021.

\bibitem{NEATReview2}
K.~O. Stanley, J.~Clune, J.~Lehman, and R.~Miikkulainen, ``Designing neural networks through neuroevolution,'' {\em Nature Machine Intelligence}, vol.~1, no.~1, pp.~24--35, 2019.

\bibitem{jaunet2021sim2realviz}
T.~Jaunet, G.~Bono, R.~Vuillemot, and C.~Wolf, ``Sim2realviz: Visualizing the sim2real gap in robot ego-pose estimation,'' {\em arXiv preprint arXiv:2109.11801}, 2021.

\bibitem{Parker2009}
G.~B. Parker, {\em Punctuated Anytime Learning to Evolve Robot Control for Area Coverage}, pp.~255--277.
\newblock Berlin, Heidelberg: Springer Berlin Heidelberg, 2009.

\bibitem{Parker2015}
G.~Parker, {\em Punctuated Anytime Learning for Autonomous Agent Control}, pp.~89--107.
\newblock Cham: Springer International Publishing, 2015.

\bibitem{parker2011fitness}
G.~Parker and J.~O'Connor, ``Fitness biasing for the box pushing task,'' in {\em 2011 IEEE International Conference on Systems, Man, and Cybernetics}, pp.~1944--1949, IEEE, 2011.

\bibitem{Steamhammer}
``Steamhammer.'' https://satirist.org/ai/starcraft/steamhammer/.
\newblock Accessed: 2024-06-17.

\end{thebibliography}
\end{document}